\documentclass[final]{cvpr}

\usepackage{times}
\usepackage{epsfig}
\usepackage{graphicx}
\usepackage{amsmath}
\usepackage{amssymb}
\usepackage[mathscr]{euscript}
\usepackage{dsfont}
\usepackage{comment}

\usepackage{multirow}
\usepackage{comment}
\usepackage{boldline}
\usepackage{amsbsy}
\usepackage{url}
\usepackage{makecell}
\usepackage{booktabs} 
\newcommand{\ra}[1]{\renewcommand{\arraystretch}{#1}} 
\usepackage{multirow} 
\usepackage{cuted} 
\usepackage{capt-of}

\usepackage{fancyhdr}

\providecommand\EN[1]{\textcolor{black}{#1}}
\providecommand\ENtwo[1]{\textcolor{black}{#1}}

\providecommand\ENthree[1]{\textcolor{black}{#1}}
\providecommand\ENfour[1]{\textcolor{black}{#1}}

\usepackage[pagebackref=true,breaklinks=true,colorlinks,bookmarks=false]{hyperref}

\def\cvprPaperID{1281} 
\def\confYear{CVPR 2021}
\begin{document}

\graphicspath{ {images/} }

\title{Body2Hands: Learning to Infer 3D Hands \\
from Conversational Gesture Body Dynamics}

\author{Evonne Ng\textsuperscript{1,2}
\hspace{0.3in} Shiry Ginosar\textsuperscript{1}
\hspace{0.3in} Trevor Darrell\textsuperscript{1}
\hspace{0.3in} Hanbyul Joo\textsuperscript{2}
\vspace{5pt}
\\
\textsuperscript{1}{UC Berkeley}
\hspace{0.3in} \textsuperscript{2}{Facebook AI Research} \\ 
}

\maketitle
\thispagestyle{fancy}
\fancyfoot{}
\chead{Appears in CVPR 2021}

\begin{strip}\centering
\vspace{-0.2in}
\includegraphics[width=\textwidth]{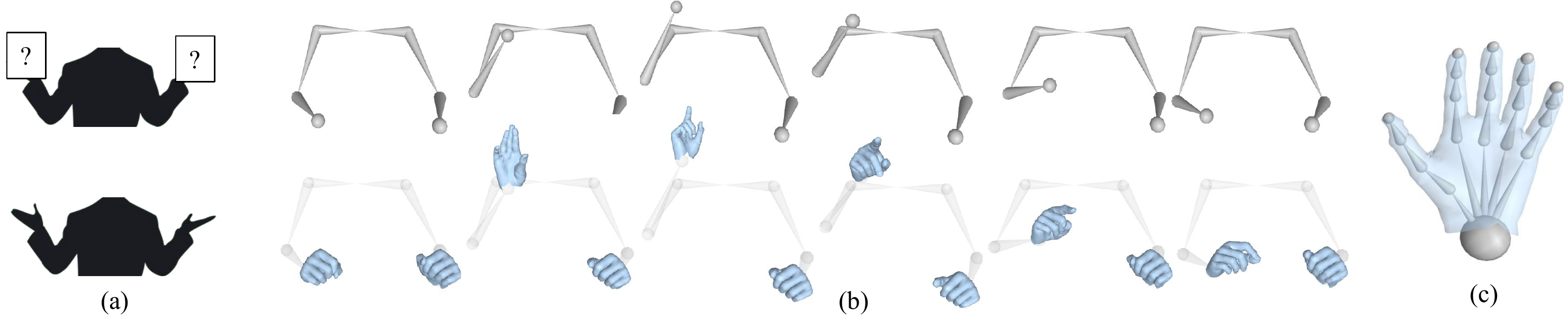}
\captionof{figure}{\textbf{Hand gestures are vital for conveying non-verbal information.} (a) Our work considers how a speaker's \emph{upper body} alone can facilitate the inference of their hand gestures. (b) From a temporal stack of the speaker's 3D body poses (top), we predict corresponding hands (bottom). (c) Body2Hands outputs a sequence of 3D hand poses in the form of an articulated 3D hand model.
Project page: \small{\url{http://people.eecs.berkeley.edu/~evonne_ng/projects/body2hands/}}
}
\label{fig:intro}
\end{strip}

\begin{abstract}
We propose a novel learned deep prior of body motion for 3D hand shape synthesis and estimation in the domain of conversational gestures. Our model builds upon the insight that body motion and hand gestures are strongly correlated in non-verbal communication settings. We formulate the learning of this prior as a prediction task of 3D hand shape over time given body motion input alone. Trained with 3D pose estimations obtained from a large-scale dataset of internet videos, our hand prediction model produces convincing 3D hand gestures given only the 3D motion of the speaker's arms as input. We demonstrate the efficacy of our method on hand gesture synthesis from body motion input, and as a strong body prior for single-view image-based 3D hand pose estimation. We demonstrate that our method outperforms previous state-of-the-art approaches and can generalize beyond the monologue-based training data to multi-person conversations.
\end{abstract}

\section{Introduction}

When we communicate,
we convey nonverbal signals with our body and hands~\cite{McNeill92}. In particular, subtle nuances can be conveyed by performing specific conversational \textit{hand} gestures, as the human hand is richly expressive with many degrees of joint freedom. This  
primordial form of communication 
is deeply ingrained in human nature. From early infancy, human babies pay extra attention to their own and others' hands~\cite{linda_smith_baby_hands} and subsequently learn to convey their needs via hand and finger gestures long before they speak. Endowing machines with the ability to perceive and use conversational hand gestures is therefore an essential step towards teaching them to effectively interact with humans. 

However, learning the intricacies of conversational hand gestures requires vast amounts of data. 
While previous approaches attempted rule-based~\cite{cassell1994animated} and data-driven~\cite{jorg2012data} methods, a \emph{learning based method} from large swaths of data would allow for both modeling
the fine-grained details of hand motion as well as generalization beyond the training set. Unfortunately, there are many challenges in capturing conversational hand gestures in realistic settings. These include the elaborate motions of fingers, the relatively small size of the hand with respect to the body, and frequent self occlusions. Such challenges make capturing the motion of human hands difficult, even for industry-level multi-camera, optical-marker-based motion capture systems~\cite{han2018online, leetalking}. Embodied 3D hand motion capture datasets in realistic conversational scenarios are therefore extremely rare. 

In this paper, we propose an approach for learning conversational hand gestures on a large-scale dataset of in-the-wild videos with noisy pseudo-ground truth.
We build upon the insight that body motion and hand gestures are strongly coupled during speech~\cite{jorg2012data}. 
By learning this correlation, we can build a reliable prior for hand gestures conditioned only on the observation of body motion.
This approach allows us to take advantage of readily available body motion data from current motion capture systems as well as single-view image-based 3D body pose estimation approaches~\cite{kanazawa2018end,xiang2019monocular}. Specifically, we formulate a 3D hand-gesture prediction problem from 3D arm motion input and demonstrate that body-hand correlations can be learned from a large-scale publicly-available monologue \textit{video} dataset.

Leveraging the learned body-motion-to-hand correlation, we present two applications: First, we propose a learned approach for realistic conversational hand gesture synthesis from body-only input (See Fig.~\ref{fig:intro} for an example). Second, we use the learned correlation as a body-motion prior for single-view 3D hand pose estimation. 
While body priors for pose estimation have been classically considered in a non-learning, general setting~\cite{raquel}, recent 3D hand trackers~\cite{zimmermann2017learning, xiang2019monocular,zhou2020monocular} have overlooked them. 

Our novelty is in proposing a \emph{learned} deep body prior for the \emph{domain-specific setting} of conversational gestures.  Focusing on a single domain of motion, such as nonverbal communication, allows us to learn a stronger prior than in the general setting. 
This prior is especially effective in scenarios where the captured appearance of the hands in video is degraded due to occlusions, motion blur, or low resolution. We demonstrate that our proposed model, trained without any clean mocap ground truth, \textit{generalizes} beyond the training set of in-the-wild monologue data to other speakers as well as to multi-person 
settings.


\begin{figure*}[t]
    \begin{center}
        \includegraphics[width=1.0\linewidth]{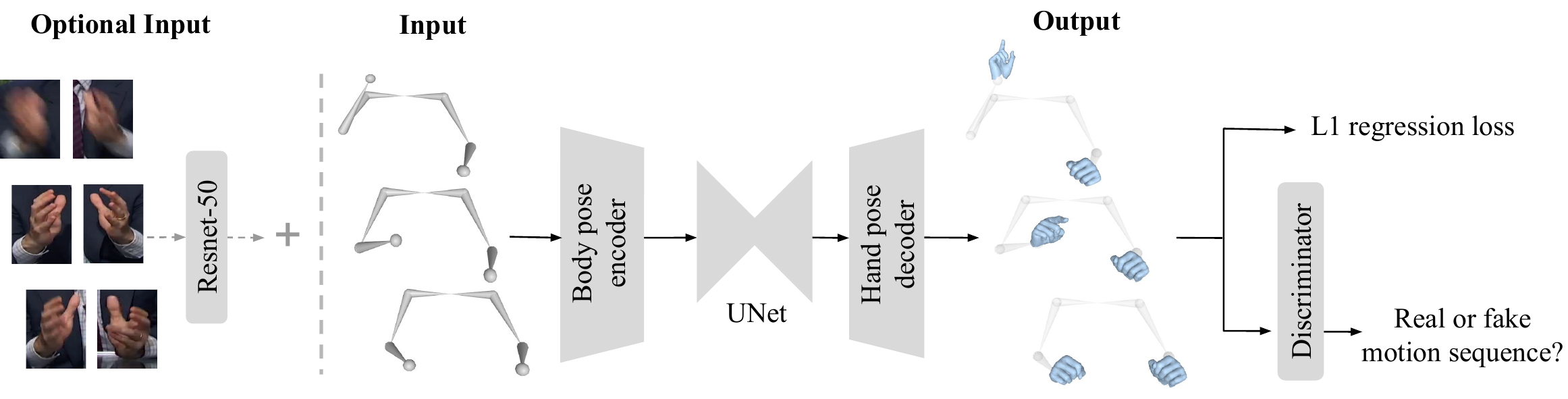}
        \caption{Encoder-Decoder network architecture for hand gesture prediction from body pose input. Our network takes a 3D body pose sequence as input. \ENfour{Optionally, our network can take an additional hand image feature (if available) as input to perform 3D hand pose estimation of the seen hands.} The body pose encoder learns inter-joint relationships, while the UNet summarizes the sequence into a body dynamics representation. Finally, the hand decoder learns a mapping from body dynamics to hands. The output is a predicted corresponding gestural hand pose sequence. L1 regression to the ground truth hand poses provides a training signal, while an adversarial discriminator ensures the predicted motion is realistic.}
        \label{fig:approach}
    \end{center}
    \vspace{-0.15in}
\end{figure*}
\section{Related Work}
\noindent \textbf{Conversational Gestures.}
Early work on generating plausible gestures for conversational agents \cite{cassell2000embodied, cassell1994animated, kopp2006towards, lee2006nonverbal, kipp2007towards} derive arm and hand motion via manually defined rules drawn from a set of annotated motion segments. Recent approaches learn person-specific gestures from speech \cite{wagner2014gesture, levine2009real, levine2010gesture, ginosar2019learning, chiu2015predicting, alexanderson2020style}, text \cite{cassell2004beat}, or 
both \cite{marsella2013virtual,yoon2019robots} without requiring hand-specified rules. However, these approaches use lab-recorded audio, text, and motion capture sequences from constrained environments \cite{levine2009real, levine2010gesture, chiu2015predicting}, which limit the variety of captured gestures, or rely on simplifying assumptions for motion generation \cite{chiu2011train, levine2010gesture, marsella2013virtual}, which cannot be generalized to in-the-wild video analysis, like ours.

More recently, \cite{ginosar2019learning} 
uses a GAN to learn a person-specific mapping from speech to 2D upper-body keypoints from in-the-wild videos. In contrast, our method focuses on the link between a speaker's body and their hands. Furthermore, we use a \emph{3D} representation,
which allows us to standardize across skeletons and hence, leverage a vast dataset of in-the-wild videos containing various speakers and settings. This enables learning gestural patterns in both person-specific and \emph{population-level} domains.

\medskip
\noindent \textbf{3D Hand Synthesis.}
Prior physics-based 3D hand synthesis approaches \cite{liu2008synthesis, pollard2005physically, zhao2013robust} generate motion patterns based on kinematic and task-specific constraints. These methods are effective when task-specific and external contact constraints are well-defined, but cannot be immediately applied in-the-wild. 
Recent data-driven approaches \cite{majkowska2006automatic, jorg2012data, mousas2015finger, stone2004speaking} define objective functions using a combination of cost terms, such as smoothness of motion \cite{jorg2012data} or transition likelihood between segments \cite{majkowska2006automatic}, to compose a trajectory of gesture sequences from a predefined collection of segmented gesture phases. Whereas the expressiveness of these methods is limited by the 
sequences contained in the collection, our approach directly regresses articulated 3D hand poses, allowing it to generalize to novel gestures and avoid the need for a predefined library. 

Probabilistic generative models \cite{levine2009real, mousas2015finger, leetalking} have also been popular in hand synthesis. Using a temporal neural network trained on an inverse kinematic loss function, \cite{leetalking} predicts hand poses from acoustic features, upper-body joint angles and velocities captured via mocap. While \cite{leetalking}, requires accurate upper-body mocap data, our approach 
trains only on in-the-wild data, which affords a richer variety of conversational situations compared to those of 
lab-constrained datasets. 
We further differentiate our approach from all prior hand synthesis approaches in that we are the first to learn a prior that can be used not only for synthesis but also to improve image-based hand pose estimation.

\medskip
\noindent \textbf{Single Image 3D Human Pose Estimation.}
Classically, 3D human pose estimation relied on information such as stereo vision~\cite{Demirdjian023-darticulated}. More recent work commonly uses 3D deformable models \cite{joo2018total, romero2017embodied, pavlakos19expressive, xu2020ghum} for markerless 3D full-body pose estimation, in which the optimized \cite{joo2018total, xiang2019monocular, pavlakos19expressive, xu2020ghum} or regressed \cite{rong2020frankmocap, choutas2020monocular, weinzaepfel2020dope, jin2020whole} reconstruction outputs are restricted to the parametric shape defined by the model. 
Building off 3D deformable models, many works have additionally exploited human pose priors \cite{kanazawa2018end,yang20183d} via data-driven approaches or explored placing constraints on joint angles or bone length \cite{zhou2017towards, dabral2018learning} to recover accurate body mesh representations from a single in-the wild image,  
but they often disregard hands \cite{kanazawa2018end,kolotouros2019spin}. Our approach leverages the success of these methods to infer the missing hands. 

Current methods in 3D hand pose estimation often focus solely on priors from the hand itself by using shape priors in deep convolutional networks to constrain the hand geometry \cite{boukhayma20193d} and 3D articulation of joints \cite{zimmermann2017learning}, or by using a structured inverse kinematics representation \cite{zhou2020monocular} to regress joint locations or mesh parameters from a single RGB image \cite{zhang2019end, hasson2019learning}. In contrast, our method leverages contextual information by explicitly capitalizing on the readily available body dynamics of the speaker. In addition to implicitly using hand shape as a prior, we emphasize the significance of using
learned human \emph{body dynamics} as a prior. More importantly, while these methods rely on clean hand images 
as input, our method can hallucinate appropriate hand poses despite heavily obstructed views-- or even, no view-- of the hands by reasoning about movements from other visible upper body parts.



\section{\ENfour{Learning Gestural Body-hand Dynamics}}\label{sec:auto-encoder}

\ENfour{Our main insight builds upon the idea that a speaker's hand gestures are highly correlated with their body motion~\cite{jorg2012data}. This enables us to leverage body motion as a strong prior for synthesizing and reconstructing 
hand motion. To learn this prior in a data-driven way, we formulate a predictive task: given the body poses of a speaker, the goal is to predict their corresponding hand poses.}


\medskip
\noindent \textbf{Problem Definition.} 
The objective of our model, $\mathscr{G}$, is to predict a sequence of 3D hand poses $\mathbf{H}$ from a corresponding sequence of 3D body poses $\mathbf{B}$:
\begin{equation}
\label{eq:hand_prediction}
\mathbf{H} = \mathscr{G}(\mathbf{B}),
\end{equation}
where $\mathbf{B}=\{ \mathbf{b}_t \}_{t=i}^T $ and $\mathbf{H}=\{ \mathbf{h}_t \}_{t=i}^T $. The 3D body pose at time $t$, $\mathbf{b}_t \in \mathds{R} ^ {18}$, is defined by the 6 3D joints of a speaker's arms (elbows, and shoulders), where 
each joint is represented by a 3D axis-angle representation in a fixed kinematic structure.
Similarly, the 3D hand pose at time $t$, $\mathbf{h}_t \in \mathds{R} ^ {126}$, is defined by 21 3D joint angles for each hand (20 finger joints and 1 global wrist orientation, as shown in Fig.~\ref{fig:intro}(c)) via 
an axis-angle representation in a fixed kinematic structure ($126=2\times 21\times 3$). 
Note that 
our model only considers 3D motion cues of 
the body and hands. 
In practice, such data is obtained by reconstructing the 3D body and hand poses from a 
parametric human model~\cite{Loper2015,romero2017embodied,pavlakos19expressive,joo2018total}, and removing the shape variation and camera parameters. In contrast to 2D representations of gesture motion~\cite{ginosar2019learning}, our normalized 3D representation 
is invariant to changes in body appearance and camera pose, allowing our model to generalize to unseen humans and scenes. 

\medskip
\noindent \textbf{An Encoder-Decoder Model for $\mathscr{G}$.} 
We use a fully convolutional 1D encoder-decoder network for our hand gesture prediction model $\mathscr{G}$.  
The network consists of a body encoder, a UNet dynamics encoder, and a hand decoder, as overviewed in Fig.~\ref{fig:approach}. The body encoder  
implicitly learns inter-joint relationships  
from an input sequence of $T$ 3D body joint rotations $\mathbf{B} \in \mathds{R}^{T \times 18}$ and outputs a body embedding $ {\boldsymbol{\phi}}_\mathbf{B} \in \mathds{R}^{T \times P}$, where $P$ is the body embedding size.
To learn a \emph{dynamics embedding} summarizing the speaker's body poses for the whole sequence, we feed the body embedding through a UNet architecture~\cite{ronneberger2015u} with a bottleneck that allows the network to pool information from past and future contexts, and skip connections
that allow high-frequency temporal information to flow through, capturing fast motions.
The UNet's temporal bottleneck of size $T'$ 
encodes the body dynamics. The output of the UNet (in $\mathds{R}^{T \times D}$, where $D$ is the dynamics embedding size), is then fed as input to the hand decoder, which learns to regress to the ground truth hand poses $\hat{\mathbf{H}} \in \mathds{R}^{T \times 126}$ via an L1 
loss:
\begin{equation}
    \mathscr{L}_{L1}(\mathscr{G}) = \|\hat{\mathbf{H}} - \mathscr{G}(\mathbf{B})\|_{1}.    
\end{equation}

\medskip
\noindent \textbf{Learning Realistic Motion Dynamics.} \label{sec:gan}
To avoid 
blending together different modes of motion, we 
introduce an adversarial discriminator~\cite{goodfellow_gan}, $\mathscr{D}$, conditioned on the temporal deltas of the predicted hand pose sequence. The discriminator ensures we produce life-like hand motion as output and facilitates 
learning 
realistic gesture dynamics. 
We use $\Delta$ to denote a function that takes as input a sequence of hand poses and outputs the difference between consecutive poses (e.g.~$[h_2-h_1, ... h_T - h_{T-1}]$). The discriminator $\mathscr{D}$ maximizes the following objective while the generator $\mathscr{G}$ minimizes it:
\begin{equation}
    \mathscr{L}_{GAN}(\mathscr{G}, \mathscr{D}) = \mathds{E}_{\mathbf{H}}[\log \mathscr{D}(\Delta(\mathbf{H})] + \mathds{E}_\mathbf{B}[\log (1-\Delta(\mathscr{G}(\mathbf{B}))].
\end{equation}
Thus, the discriminator learns to classify real or fake motion dynamics, inherently nudging the generator to produce more realistic speaker hand movements. Our full objective is thus:
\begin{equation}
    \min_{\mathscr{G}} \max_{\mathscr{D}} \mathscr{L}_{GAN} (\mathscr{G,D}) + \lambda \mathscr{L}_{L_1}(\mathscr{G}).
\end{equation}
Figure \ref{fig:approach} overviews the model; see Supp.~for architecture details.


Our hand gesture prediction model 
implicitly learns 
the interplay between a speaker's body and their 
hands, which 
can be used as a strong prior 
to  
\ENfour{synthesize} realistic communicative hand gestures.


\medskip
\noindent \textbf{3D Hand Pose Estimation with Body Pose Priors.} \label{para:estimation}
We can extend our model\ENfour{, incorporating it as a novel prior for} hand pose estimation by using an additional hand image when such input is available: 
\begin{equation}
\label{eq:hand_prediction_with_image}
\mathbf{H} = \mathscr{K}( \mathbf{I}_h, \mathbf{B}),
\end{equation}
where $\mathbf{I}_h$ is a series of hand images cropped around the left and right hand regions of the input video. 
\ENtwo{
Dropping the last fully connected layer of a ResNet-50 model~\cite{he2016deep} pretrained on ImageNet, we treat the rest of the ResNet-50 as a fixed feature extractor for $\mathbf{I}_h$. The resulting appearance-based feature is passed through a linear layer, which outputs the image embedding ${\boldsymbol{\phi}}_\mathbf{I} \in \mathbf{R}^{T \times Q}$, where $Q$ is the image embedding size. The image embedding is then concatenated with the body embedding ${\boldsymbol{\phi}}_\mathbf{B}$ and fed through the remaining unchanged UNet 
and hand decoder pipeline (See Fig.~\ref{fig:approach}). 
}

While existing single-view 3D hand pose estimation approaches~\cite{zimmermann2017learning, zhou2020monocular, zhang2019end, hasson2019learning} rely only on hand images as input, our method also leverages the inter-correlation between body pose and hand gestures. 
Thus, our prior based on body motion 
provides an additional cue for hand gesture estimation to overcome challenges caused by fundamental depth ambiguity, frequent self-occlusion, and severe motion blur. Furthermore, we consider the temporal aspect of the input, allowing our method to produce smoother, more realistic hand sequences.


\section{Experiments}

We evaluate our learned body prior \ENfour{from two angles.} 
First, we evaluate the quality of our \emph{3D hand synthesis} method by performing a perceptual user study on results from in-the-wild videos. Second, we quantitatively compare our body prior model in \emph{single-view 3D hand pose estimation from video} against 
state-of-the art hand pose estimation methods~\cite{xiang2019monocular, zhou2020monocular} on a ground-truth dataset captured in the Panoptic Studio \cite{joo2015panoptic}.
In all experiments, our method is trained and tested in a person-agnostic setting: we train and test on many different speakers. To test the generalizability of our method, we ensure that speakers appearing in the test clips do \emph{not} appear in the training set. We use 3D body pose estimates from MTC~\cite{xiang2019monocular} as input to our model during testing.

\subsection{Datasets}

To study common body and hand gesture dynamics, we produce 3D body and hand  
annotations to accompany the large scale, in-the-wild monologue dataset presented in~\cite{ginosar2019learning}.
For evaluation, we utilize a smaller multi-person conversation dataset with clean 3D body and hand ground truth captured in a multi-view setting via the Panoptic Studio~\cite{joo2015panoptic}.

\medskip
\noindent \textbf{In-the-wild 3D Body and Hand Gesture Dataset.}
Learning a body pose prior for hand gesture prediction 
requires a large-scale dataset capturing natural 3D body and finger movements that occur simultaneously with speech.
However, such motion capture datasets are extremely rare due to challenges in capturing hands. We thus leverage a state-of-the-art monocular 3D pose estimation algorithm, MTC~\cite{xiang2019monocular}, to reconstruct 3D body and hands from a large scale collection of public monologue videos~\cite{ginosar2019learning}. MTC estimates 3D body and hands in the form of a 3D parametric human model (Adam~\cite{joo2018total}), parameterized by shape parameters and 62 3D joint rotation pose parameters (22 joints for body and 20 joint for each hand). From MTC's output, we use the pose parameters corresponding to the arms and fingers as pseudo ground-truth to train our models (Eq.~\ref{eq:hand_prediction} and Eq.~\ref{eq:hand_prediction_with_image}). Although the reconstruction outputs from MTC contain failure cases with noticeable motion jitter and artifacts, we found them  
\EN{adequate for training our model when used on video with sufficient resolution}. 
To cover a broad range of gesture styles, we annotate 81 hours of in-the-wild videos for 8 gesturing speakers covering a wide range of topics from a variety of different settings (e.g. television shows, lectures, religious sermons).

\begin{figure*}[t]
    \begin{center}
        \includegraphics[width=1.0\linewidth]{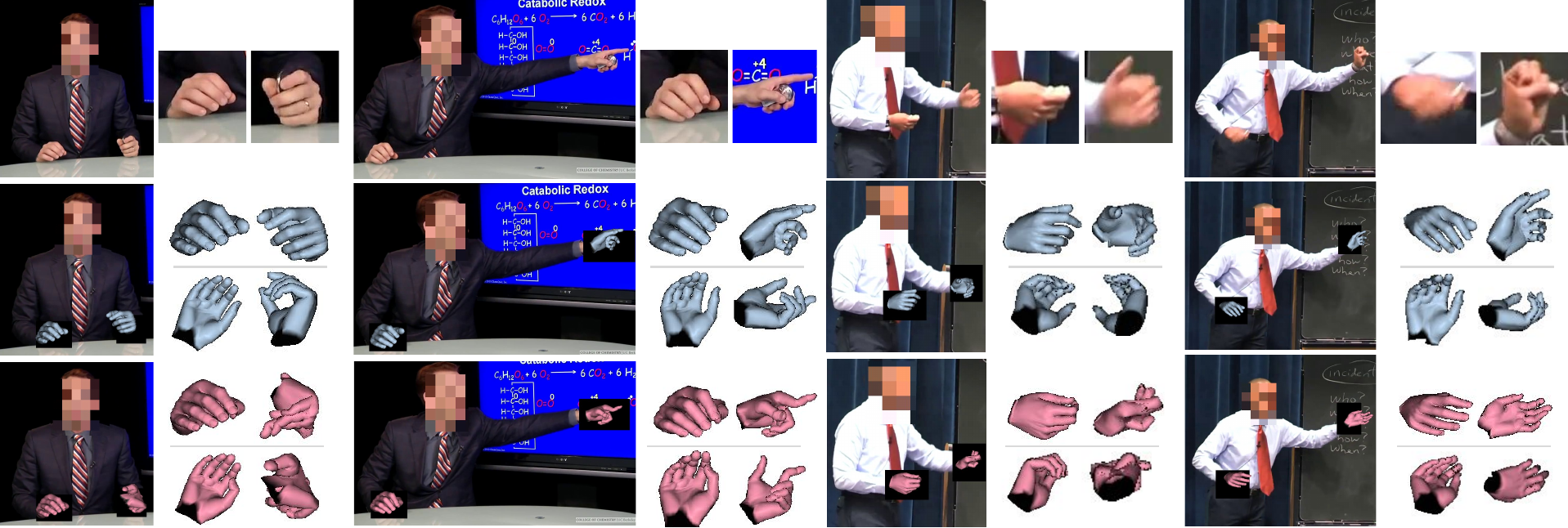}
        \caption{
        Our predicted 3D hand poses against a SOTA image-based method, MTC~ \cite{xiang2019monocular}. We show each prediction from a novel view below their respective hand. 
        Row 1: View of speaker and magnified hands (not used by our method). Row 2: results from our method, using body-only as input. Row 3: MTC \cite{xiang2019monocular} image-based results. We show results from a novel video of a person in the training set (left) and a novel video of an unseen person (right) to demonstrate that our model generalizes across individuals. Note: all evaluations discussed in the paper are only on unseen people. 
        For motion blurred or self occluded hands, ours produces more accurate results.
        }
        \label{fig:success}
    \end{center}
    \vspace{-0.10in}
\end{figure*}

\medskip
\noindent \textbf{Panoptic Studio Dataset.}
Given the lack of clean ground truth for the in-the-wild capture, we utilize an additional Panoptic Studio dataset presented in \cite{joo2019towards} for evaluation purposes. This dataset consists of 120 sequences (about 2 hours) of individuals naturally conversing in a negotiation game scenario. To handle 
severe occlusions among 
other individuals in the scene,
the dataset uses a multi-view approach (with 31 HD cameras) to reconstruct 3D body and hand keypoint locations. We apply an additional parametric 3D human model reconstruction algorithm~\cite{joo2018total} to convert the keypoints to a 3D angle representation, 
matching the 3D human model form (Adam) of our in-the-wild dataset.
\ENthree{All data and annotations will be publicly released at \url{http://people.eecs.berkeley.edu/~evonne_ng/projects/body2hands/.}}


\subsection{Implementation Details}

We generate training data by creating sliding windows of size 64 with an overlap of 32 frames for each sequence of the in-the-wild training set. This yields 139K total sequences, for which we use a 70/30 training/validation split. For the encoder-decoder model, we use embedding dimensions $P=D=Q=256$. We train the model using Adam with a batch size of 128 and a learning rate of $10^{-4}$. We train with an adversarial loss every third epoch, and without, for all other epochs. 
Training time is 2 hours on a single GeForce RTX 2080 GPU for 200 epochs.
See Supp.~for additional details.


\subsection{3D Hand Synthesis}
\ENfour{
We assess whether our approach generates realistic and perceptually convincing hand pose sequences for the speaker. With \emph{only} 3D body pose annotations as input, our model (defined in Eq.~\ref{eq:hand_prediction}) hallucinates a 
corresponding hand gesture sequence. This model can be particularly useful for synthesizing realistic hand gestures for body-only data, where hand motion capture is unavailable. For example, body-only mocap data (e.g., CMU Mocap~\cite{cmu-graphics} and KIT~\cite{mandery2015kit}) or existing 3D body pose estimation methods~\cite{kanazawa2018end, kolotouros2019spin} that output torso and limb poses only. See Supp.~video for examples where our method synthesizes hands given body-only mocap or off-the-shelf 3D body pose estimates as input.} 

\medskip
\noindent \textbf{Perceptual Evaluation.} 
We design a perceptual evaluation to corroborate our quantitative results.
The \ENthree{perceptual} evaluation compares our synthesis results (without using any hand observation), \textbf{Ours w/ $B$}, against MTC~\cite{xiang2019monocular}. 
Comparing against state-of-the-art in 3D hand pose estimation allows us to qualitatively evaluate how close results from using \emph{the body prior only (ours)} can get to results from using the true hand image.
\ENthree{Furthermore, the perceptual study ensures our quantitative metrics are well founded.}

For the evaluation, we collect a new corpus of 12 in-the-wild YouTube videos from 9 speakers \emph{not seen} in the training set. We visualize the hands by reembodying the predicted hands on the corresponding MTC-extracted body pose. The reembodied hands are then overlayed on the corresponding video frame with a black box over the true hands of the speaker. 
To provide evaluators with additional perceptual context, we also include audio. See Fig.~\ref{fig:success} and supp. video for example visualizations.

Evaluators watched a series of video pairs (shown synced side by side), in which the hands of one video are synthesized by our method, and the other, are estimated by MTC. \ENthree{The left/right location of ours vs. MTC was randomized for each video.} 
Evaluators were then asked to ``vote" for which hand sequence looks more realistic on the given speaker. They were given unlimited time to answer and could replay the video. We randomly sampled 19 pairs of 12-second clips, each of resolution $1280 \times 720$, from the YouTube corpus. Our results were scored by 45 evaluators.

\begin{table*}[t]
    \centering
    \ra{1.3} 
    \begin{tabular}{@{}lccccccc@{}}
    \toprule
    & Ours w/ $B+I$ & Ours w/ $B$ & MTC \cite{xiang2019monocular} & FrankMocap \cite{rong2020frankmocap} & Minimal \cite{zhou2020monocular} & NN & Median \\
    \midrule
    
    unclear & \textbf{2.53 $\pm$ 0.8} & 2.85 $\pm$ 0.5 & 5.21 $\pm$ 0.9 & 4.89 $\pm$ 0.4 & 9.30 $\pm$ 0.9 & 4.32 $\pm$ 0.7 & 4.78 $\pm$ 0.6 \\
    
    clear & 3.33 $\pm$ 0.6 & 3.45 $\pm$ 0.6 & \textbf{2.91 $\pm$ 1.4} & 3.02 $\pm$ 0.6 & 5.46 $\pm$ 1.6 & 4.51 $\pm$ 0.4 & 4.61 $\pm$ 1.0 \\
    
    all frames & 2.81 $\pm$ 0.8 & 3.05 $\pm$ 0.6 & 3.66 $\pm$ 1.3 & 3.59 $\pm$ 0.6 & 7.30 $\pm$ 1.7 & 4.42 $\pm$ 0.7 & 4.70 $\pm$ 0.7 \\
    \bottomrule
    \end{tabular}
    \vspace{0.05in}
    \caption{Avg.~joint errors with std.~(in mm; approximated by 30 cm avg.~shoulder length; lower is better) 
    on the Panoptic Studio capture, broken down for frames with clear/unclear view of hands. While MTC and FrankMocap expectedly outperform 
    on clear views of the hands, the margin separating ours vs. MTC and ours vs. FrankMocap 
    is small. Our method using body priors (with or without image input) outperforms all baselines by a larger margin when 
    hands are unclear (low-res, obstructed). Avg.~error over all frames also shown.}
    \label{tab:baselines}
\end{table*}

\begin{figure*}
    \begin{center}
        \includegraphics[width=1.0\linewidth]{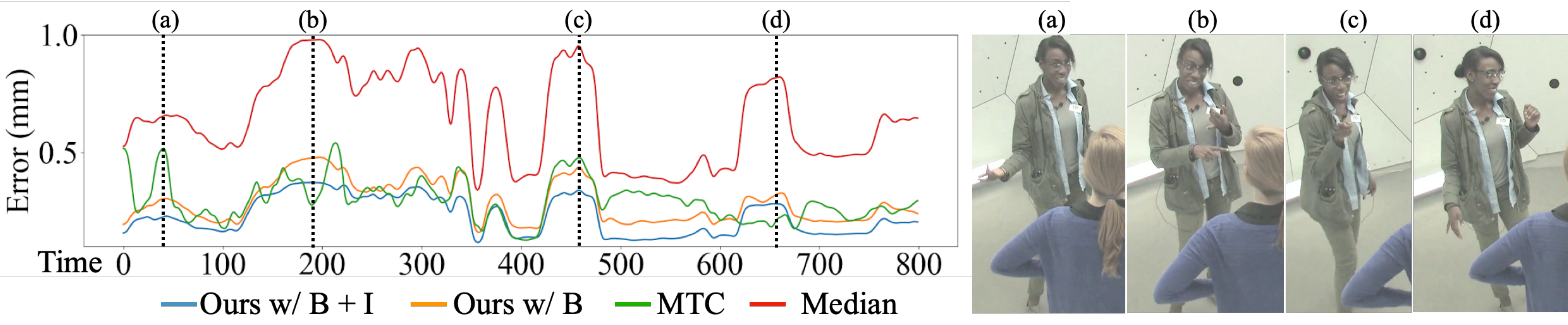}
        \caption{Analysis of typical errors. Error over time plotted on the left (lower is better). Frames shown for notable scenarios on the right.  MTC fails from (a) naturally arising occlusions or from (c) motion blur/low resolution on hands. With clear views of the hands (b) and (d), MTC performs slightly better, though the margin separating ours from MTC is smaller than in cases where MTC fails. Overall, ours outperforms other baselines whether we take as input an image observation or not.}
        \label{fig:graph}
    \end{center}
    \vspace{-0.15in}
\end{figure*}

\medskip
\noindent \textbf{Synthesis Results.}
%
Our method, based purely on a body motion prior, is qualitatively competitive against current image-based state-of-the-art in 3D hand pose estimation~\cite{xiang2019monocular}. 
\EN{In total, 54.54\% of votes indicated that the synthesized hands from \textbf{Ours w/ $B$} }
\ENthree{are perceptually competitive against \emph{real hand gestures} approximated by \textbf{MTC}.}
This illustrates the surprising result that \emph{even without seeing the hands}, our method can synthesize hands from body-input alone that people find 
\ENthree{as realistic as} those synthesized by a method that uses the true hand image to estimate the actual shape of the hand.

\ENtwo{In Figure \ref{fig:success}, \textbf{Ours w/ $B$} outperforms \textbf{MTC} in predicting hands that are self-occluded or blurry, better capturing a clasped hand or a hand swinging up to point.}
\ENtwo{The lowest 
votes our method receives on a video is $36\%$, and highest, $77\%$.} We note that in videos where \textbf{MTC} outperforms, the speaker's hand pose is often semantically linked to auditory keywords (e.g.~holding up the index finger and saying ``one"), which may be missed by observing only the body pose of the speaker. See Supp.~for details and additional perceptual evaluations. 

Our method successfully synthesizes perceptually convincing hand poses---without any pixel information---by observing the body pose of the speaker. This supports our key technical insight that hands are closely linked to body dynamics via a \emph{learnable} relation. We can therefore explicitly leverage body pose as an effective prior to improve hand pose estimation.

\begin{figure*}[t]
    \vspace{-0.15in}
    \begin{center}
        \includegraphics[width=1.0\linewidth]{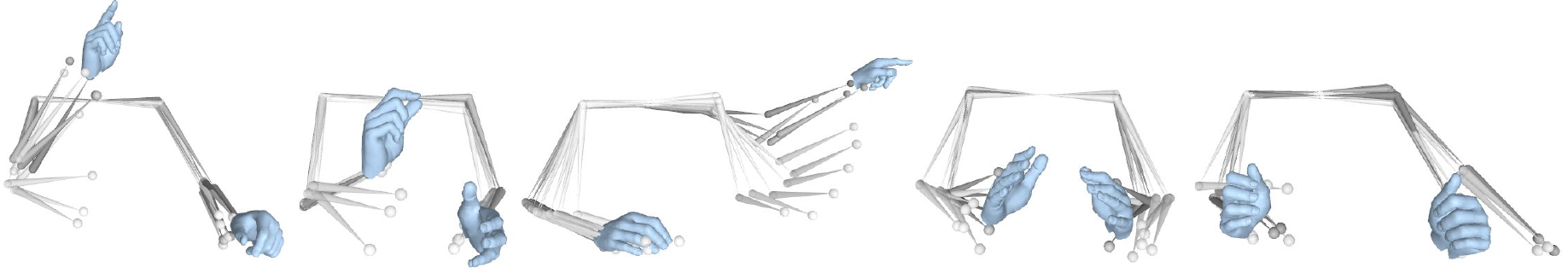}
        \caption{
         For each hand pose example query, we find 10 closest predicted hand poses from in-the-wild videos and visualize their corresponding body poses (darker means closer match). We reembody the query hands on its corresponding body shown in darkest shade. Body2Hands captures distinct correlations between the body and hands for common communicative gestures.
        }
        \label{fig:priors}
    \end{center}
\end{figure*}

\begin{figure*}[t]
    \begin{center}
        \includegraphics[width=1.0\linewidth]{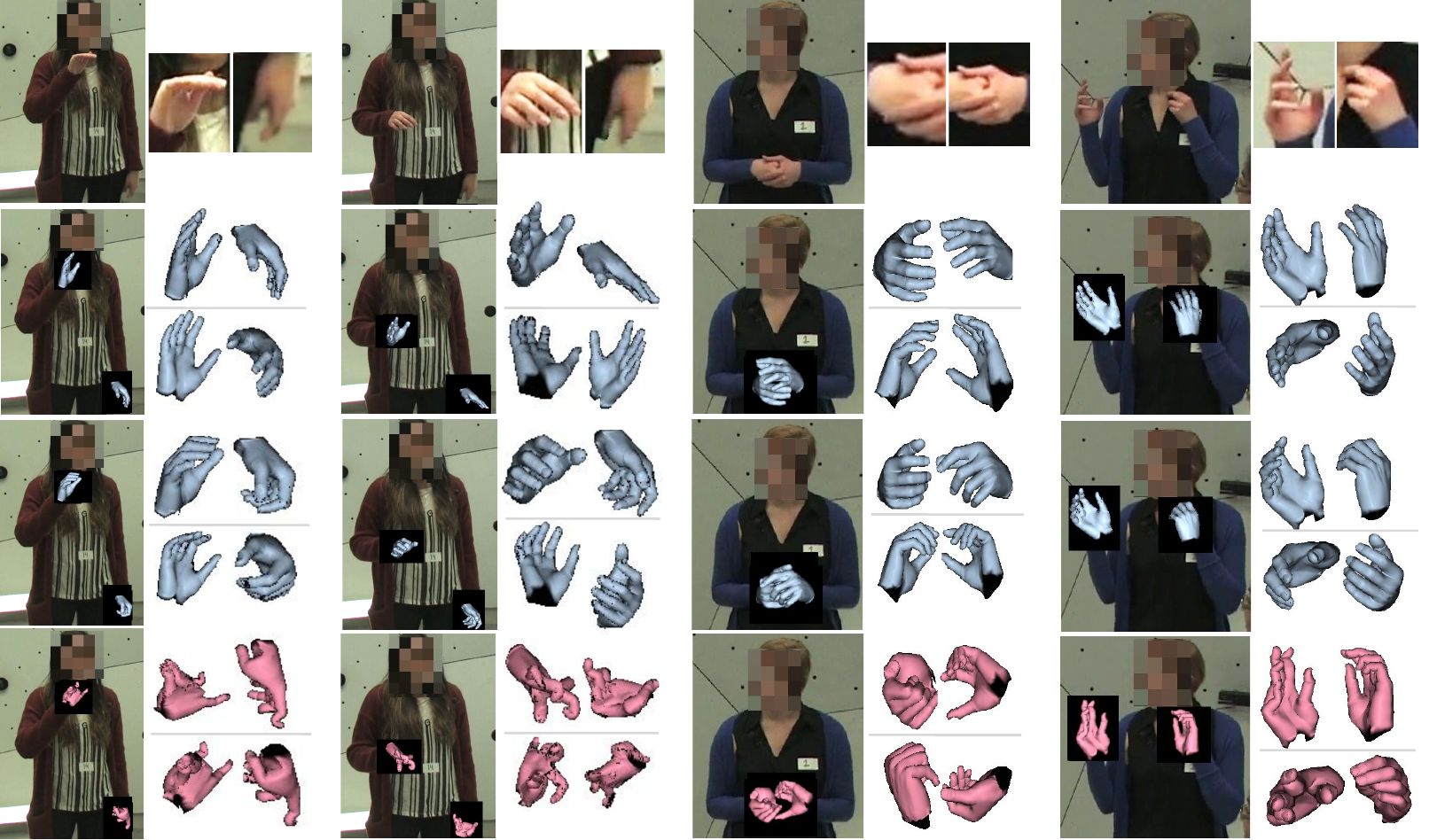}
        \caption{Adding the image embedding improves the accuracy for examples where using the body prior alone fails due to low/no distinguishable body motion. We show two views for each prediction below their respective hands. Row 1: View of speaker and magnified hands. Row 2: Our results  from using body prior alone. Row 3: Adding an image feature with the prior addresses this issue, snapping to the true hands. Row 4: MTC~\cite{xiang2019monocular} image-based results. Shown are ``unclear" frames as referred to in Tab~\ref{tab:baselines}. See video for all frames. }
        \label{fig:failure}
    \end{center}
    \vspace{-0.15in}
\end{figure*}

\begin{table*}[t]\footnotesize
    \centering
    \ra{1.3} 
    \begin{tabular}{@{}lccccccc@{}}
    \toprule
    & Ours w/ $B+I$ & Ours w/ $B$ & Ours w/ $I$ & Ours w/ GT $B+I$ & Ours w/ GT $B$ & Ours w/ $B+I$ no GAN & Ours w/ $B$ no GAN \\
    \midrule
    all frames & 2.81 $\pm$ 0.8 & 3.05 $\pm$ 0.6 & 9.40 $\pm$ 0.6 & 2.45 $\pm$ 0.07 & 2.49 $\pm$ 0.05 & 2.84 $\pm$ 0.5 & 3.19 $\pm$ 0.4 \\
    \bottomrule
    \end{tabular}
    \vspace{0.05in}
    \caption{Effect of the body pose prior and discriminator on our method. Avg.~joint errors with std.~(in mm; approx.~by 30 cm avg.~shoulder length; 
    lower is better). Evaluated on Panoptic Studio capture. Performance worsens without the body prior (Ours w/ $I$), and improves with more accurate ground truth body pose inputs. Discriminator results in only small quantitative differences.} 
    \label{tab:ablation}
    \vspace{-0.10in}
\end{table*}

\subsection{Single-image 3D Hand Pose Estimation}

We quantitatively assess the advantage of our body pose prior for 3D hand pose estimation.
We use the Panoptic Studio capture with ground-truth data to
compare  
against alternative approaches, including state-of-the-art in monocular hand motion capture, \textbf{MTC}~\cite{xiang2019monocular}, \textbf{Minimal Hand}~\cite{zhou2020monocular}, and \textbf{FrankMocap}~\cite{rong2020frankmocap}. 
In this experiment, we use the same monocular video for all methods by selecting the camera view with the clearest, non-occluded frontal view of the speaker for each game. Yet, the social game scenario (three individuals in each game) is challenging for monocular hand motion capture approaches since (1) individuals are far from the cameras, making the image resolution of hands low, and (2) occlusions naturally and frequently emerge during conversations. 
We demonstrate that our body prior model is effective in this challenging 
scenario. We compare the following methods:

\vspace{-0.05in}

\begin{itemize}
    \item \emph{Ours with body (\textbf{Ours w/ $B$}):} \ENthree{Our synthesis} method \ENfour{from the perceptual evaluation} 
    (Eq.~\ref{eq:hand_prediction}). Takes only body pose as input. 
    \vspace{-0.05in}
    \item \emph{Ours with body and image (\textbf{Ours w/ $B$ + $I$}):} Our hand pose estimation model presented in Eq.~\ref{eq:hand_prediction_with_image}. Takes both body pose and image as input. Please note, the image feature we use here is weaker than those used by previous state-of-the-art methods, which are trained specifically on 2D \cite{simon2017hand} or 3D \cite{xiang2019monocular, zimmermann2019freihand, hampali2020honnotate} hand ground truth. In contrast, we opted for a naive ResNet image feature as an additional input.
    
    \vspace{-0.05in}
    \item \emph{Monocular Total Capture  (\textbf{MTC}) \cite{xiang2019monocular}:} State-of-the-art 
    full body + hand 3D pose estimation, based on an optimization framework.
    
    \vspace{-0.05in}
    \item \emph{FrankMocap (\textbf{FrankMocap}) \cite{rong2020frankmocap}:} State-of-the-art  
    full body + hand 3D pose estimation, based on networks that regress 3D pose parameters.
    
    \vspace{-0.05in}
    \item \emph{Minimal Hand (\textbf{Minimal}) \cite{zhou2020monocular}:}
    Current state-of-the-art monocular 3D hand pose estimation. For each video frame, we extract cropped bounding box images of both the left and right hands using OpenPose. We feed both images individually through the pretrained model provided by the authors to extract 3D keypoint estimations of each hand.
    
    \vspace{-0.05in}
    \item \emph{Nearest Neighbors (\textbf{NN}):} \ENthree{A segment-search method commonly used for synthesis in graphics. We first break the input sequence into smaller sub-segments.} For each body pose \ENthree{sub-segment}, we find its nearest neighbor from the training set and transfer its corresponding hand pose \ENthree{sub-segment}. 
    
    \vspace{-0.05in}
    \item \emph{Always predict a median pose (\textbf{Median}):} A simple yet strong baseline 
    exploiting the prior that a speaker's hands are at rest most of the time \cite{kendon2004gesture}. We obtain 
    median hand pose from the test set. 
\end{itemize}

\medskip
\noindent \textbf{Evaluation Metric.}
We compute the average Euclidean distance between the predicted and ground truth 3D hand joint locations. Since the scale of 3D joints from each 3D parametric model differs (Adam~\cite{joo2018total} for ours and MTC, MANO~\cite{romero2017embodied} for Minimal Hand, and SMPL-X for FrankMocap), we apply rigid alignment by Procrustes analysis \cite{gower1975generalized} to normalize for scale and global orientation before computing errors (commonly done in the 3D pose estimation field~\cite{xiang2019monocular, kanazawa2018end}). We report average error over the entire sequence, and scale the error unit to millimeters based on a reference shoulder distance of 30 cm.

\medskip
\noindent \textbf{Estimation Results.}
Table~\ref{tab:baselines} shows our method outperforms all other competing methods under challenging scenarios. Errors are shown averaged over all 42 hand joints.

Our body-only approach \textbf{Ours w/ $B$} makes significant gains over \textbf{MTC}, \textbf{FrankMocap}, and \textbf{Minimal} on frames with difficult views of the hands ($\approx 60\%$ of frames). While \textbf{MTC} and \textbf{FrankMocap} expectedly outperform on clean views of the hands, the margin separating either from both our body prior methods is much lower. This result supports our key technical novelty in using body as a strong prior for hand pose estimation. Despite 31 camera views in the Panoptic Dome, finding a view where hands are constantly and clearly in-view is difficult \ENtwo{(shown in Figure \ref{fig:graph})}. By observing the body dynamics, our method naturally overcomes occlusion and motion blur, outperforming image-based methods. Thus, information provided by the more visible body pose is essential for deriving accurate hand pose estimates. Further, the results demonstrate our model, trained on monologue videos, generalizes to broader conversational domains.

While \textbf{Ours w/ $B$} alone outperforms all other baselines, by adding a weak appearance-based cue, \textbf{Ours w/ $B+I$} improves performance, though the margin is smaller than those separating \textbf{Ours w/ $B$} from all other baselines. This demonstrates the effectiveness of using body priors to inform image-based methods towards more accurate hand pose estimates.

Figure \ref{fig:priors} shows examples of links our method discovers between the hand and body pose of various speakers. For each chosen hand pose, we find the 10 closest predicted hand poses from the test set and visualize the body pose at that instance. More distinctive hand patterns with semantic significance (e.g. pointing or pinching to show "a little bit") have the most consolidated body poses, demonstrating a stronger link between the hands as a function of the body. While non-semantic hand poses may be associated with a greater variety of body poses, the overall composition of the body poses is still noticeably distinct. For example, the bodies accompanying the straight hands are always upright, while those of the clasped hands have a bit of a lean.

In Figure \ref{fig:failure} we show examples where \textbf{Ours w/ $B$} predicts incorrect, albeit reasonable, hand poses, while \textbf{Ours w/ $B+I$} uses the additional appearance cue to accurately snap to the correct hand pose. 

Table~\ref{tab:ablation} quantifies the impact of the body prior in our network. As demonstrated in an ablation, using only a weak appearance cue \textbf{Ours w/ $I$} in our network is not sufficient to capture the speaker's hand pose, leading to poorer results.
We also substitute in the 3D ground truth body pose of the speaker (i.e. the true pose given by the Panoptic Studio) for the extracted MTC body pose to test how the source of the body estimates affect our results in \textbf{Ours w/ GT $B+I$} and \textbf{Ours w/ GT $B$}. We see that more accurate body poses can further improve results, while improvement gains from the additional image feature is less impactful with more accurate body pose inputs. Further, Table~\ref{tab:ablation} shows that ablating the GAN by removing the discriminator in our method leads to only small quantitative differences.
\EN{See Supp.~for video, person-specific models, 
architecture details, and a study analyzing the contributions of using our body-hand prior vs. a temporal prior from using hand pose \emph{sequences}.}


\section{Discussion}
We present Body2Hands: a data-driven approach towards learning the interplay between a speaker's hand and their body dynamics. We leverage this link to exploit the more visible body pose as a strong prior for hand synthesis and estimation. Trained only on in-the-wild data, our method's 
results on two datasets across 
various individuals and conversational settings demonstrate the promise of our idea. As a limitation, our method is restricted to the domain of conversational hand gestures and fails when the speaker is holding a pen or prop. Future work includes integrating our prior with more sophisticated image-based approaches, \ENfour{making our body-pose-only synthesis framework nondeterminsitic}, incorporating other observable priors such as audio, and extending 
beyond conversational gestures.

\medskip
\noindent \textbf{Acknowledgements.} The work of Ng and Darrell is supported in part by DoD including DARPA's XAI, LwLL, Machine Common Sense and/or SemaFor programs, as well as BAIR's industrial alliance programs. Ginosar's work is funded by the NSF under Grant \# 2030859 to the Computing Research Association for the CIFellows Project.

{\small
\bibliographystyle{ieee_fullname}
\bibliography{egbib}
}

\end{document}